\documentclass[sn-mathphys,Numbered]{sn-jnl}

\usepackage{graphicx}%
\usepackage{multirow}%
\usepackage{amsmath,amssymb,amsfonts}%
\usepackage{amsthm}%
\usepackage{mathrsfs}%
\usepackage[title]{appendix}%
\usepackage{xcolor}%
\usepackage{textcomp}%
\usepackage{manyfoot}%
\usepackage{booktabs}%
\usepackage{algorithm}%
\usepackage{listings}%
\usepackage{algorithmic}
\usepackage{array}
\usepackage{enumitem}
\usepackage[font=small,skip=0pt]{caption}
\usepackage{hyperref}
\usepackage{xcolor}
\usepackage{multirow}
\usepackage{makecell}

\usepackage{tabularx}

\begin{document}

\title[Article Title]{C-LEAD: Contrastive Learning for Enhanced Adversarial Defense}









\author*[1]{\fnm{Suklav} \sur{Ghosh}}\email{suklav@iitg.ac.in}

\author[1]{\fnm{Sonal} \sur{Kumar}}\email{k.sonal@iitg.ac.in}

\author[1]{\fnm{Arijit} \sur{Sur}}\email{arijit@iitg.ac.in}


\affil[1]{\orgdiv{Department of Computer Science and Engineering}, \orgname{Indian Institute of Technology}, \orgaddress{\street{Guwahati}, \city{Guwahati}, \postcode{781039}, \state{Assam}, \country{India}}}

\abstract{Deep neural networks (DNNs) have achieved remarkable success in computer vision tasks such as image classification, segmentation, and object detection. However, they are vulnerable to adversarial attacks, which can cause incorrect predictions with small perturbations in input images. Addressing this issue is crucial for deploying robust deep-learning systems. This paper presents a novel approach that utilizes contrastive learning for adversarial defense, a previously unexplored area. Our method leverages the contrastive loss function to enhance the robustness of classification models by training them with both clean and adversarially perturbed images. By optimizing the model’s parameters alongside the perturbations, our approach enables the network to learn robust representations that are less susceptible to adversarial attacks. Experimental results show significant improvements in the model's robustness against various types of adversarial perturbations. This suggests that contrastive loss helps extract more informative and resilient features, contributing to the field of adversarial robustness in deep learning. The code is publicly made available on GitHub in the following link: \url{https://github.com/suklav/C_Lead}.
}

\keywords{Adversarial training, Contrastive learning, Representation Learning, Computer Vision, Deep Learning.}



\maketitle

\section{Introduction}\label{sec1}

\begin{figure}[h]
    \centering
    \includegraphics[scale=0.47]{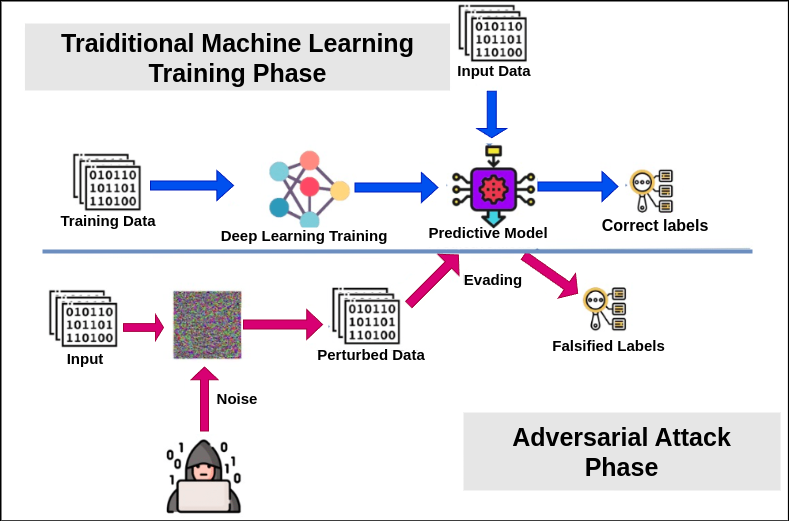}
    \caption{A general framework for the adversarial attack in a deep learning network.}
    \label{fig:common_framework_adv_attack}
\end{figure}

Deep learning is one of the most widely used tools in computer vision research. It enables us to develop deep neural networks like convolutional neural networks (CNN) and vision transformers to perform various computer vision tasks. Before deploying in practical scenarios, these models undergo crucial training and testing on extensive datasets. 
However, such models are vulnerable to attacks that manipulate predictions by introducing visually imperceptible perturbations to training images and videos, known as adversarial perturbations\cite{eleftheriadis2024adversarial,chen2024adversarial,pramanick2025trans,soor2025universal,soor2025universal1}. Hence, it's essential to take certain measurements before utilizing deep learning models for critical computer vision applications. A general framework for adversarial attack is described in Figure \ref{fig:common_framework_adv_attack}.

One major problem with using neural networks in safety-paramount applications, such as autonomous driving, has been their susceptibility to miniature perturbations \cite{SzegedyZSBEGF13}. To guarantee the trained networks' resilience towards adversarial attacks \cite{zhang2019theoretically,tramer2020ensemble,madaan2020adversarial}, random noise\cite{zheng2016improving}, and corruption\cite{hendrycks2019benchmarking,yin2019fourier}, a number of articles were put forward. In order to achieve the highest possible loss on the target framework, adversarial learning—which trains the framework using perturbed samples—may be among the most often used methods for achieving adversarial resilience. Adversarial learning has advanced significantly through recent years, beginning with the fast gradient sign method(FGSM), which employs a perturbation along the gradient direction, and moving towards projected gradient descent(PGD), which provides the highest loss throughout iterations, and TRADES, which compromises between adversarial robustness and clean accuracy \cite{goodfellow2015explaining,madry2018towards,zhang2019theoretically}. Despite this, in order to produce adversarial attacks, traditional adversarial learning strategies must have class labels.
\\
Self-supervised learning\cite{gidaris2018unsupervised,noroozi2016unsupervised,chen2020simple,he2020momentum,wu2018unsupervised} has gained prominence in the past few years as a method of learning representations for deep neural networks. It involves training the framework on unlabeled data in a supervised way using self-generated labels out of the data itself\cite{carmon2019unlabeled}. These self-supervised learning techniques include, for instance, tackling randomized Jigsaw puzzles\cite{noroozi2016unsupervised} and predicting the angle of rotation\cite{gidaris2018unsupervised}. Instance-level identity retention combined with contrastive learning has been demonstrated to be a highly successful method for acquiring rich representations for classification\cite{chen2020simple,he2020momentum}. The general goal of contrastive self-supervised learning architectures, like those found in\cite{chen2020simple,he2020momentum,wu2018unsupervised,tian2020contrastive}, is to minimize an instance's closeness to other examples while maximizing its resemblance to its augmentation. 
\\
The stated contrastive learning is a widely studied representation learning approach \cite{kumar2023globally}. A general contrastive learning framework consists of a negative and positive pair sampling strategy, a deep learning model (feature extractor), and a contrastive loss (an objective function).
The contrastive loss function minimizes the distance of positive pairs and maximizes the distance of negative pairs in the feature representation space. 
\\
In our paper, we extend the idea of pair sampling strategy and the objective function of the contrastive learning approach for adversarial defense training. 

Our modified sampling strategy creates a positive pair by sampling multiple perturbed versions of an image and a negative pair by sampling multiple perturbed versions of different images. For clarification, all images in a positive pair are perturbed versions of the same image, and each image in a negative pair is a perturbed version of different images from the dataset. The intuition is to bring the anchor image and its different perturbed versions close in a feature representation space with a contrastive learning approach. The perturbed versions of images are generated with existing adversarial attack mechanisms like FGSM, PGD, and CW. The deep learning model, pre-trained with our method, can produce a similar representation for an image and its perturb versions. Later, we utilize the robust pre-trained model to perform downstream tasks like image classification. 
It also acts as a filter for downstream tasks to prevent adversarial attacks. The experimental results show that the proposed contrastive adversarial training makes the feature extractor or the CNN backbone robust enough to handle perturbed images by producing a visual feature representation similar to the anchor images. 
 \\
The major contributions of our paper are:
\begin{enumerate}
\item Introduced a novel framework based on contrastive learning to enhance deep learning model robustness, thoroughly investigating various attack techniques and developing a resilient model capable of withstanding both known and unknown gradient-based attacks during training and testing stages.
\item Achieved significant improvements in backbone model performance through experiments: FGSM attack resistance increased by 40\%, PGD attack resistance enhanced by 53\%, and CW attack resistance improved by 41\%.
\end{enumerate}
The rest of the paper is organized as follows.
Section 2 reviews the Literature, and Section 3 presents the proposed methodology. Section 4 elucidates the results \& analysis section, which is followed by the conclusion section.

\section{Related Work}\label{sec2}

\subsection{Contrastive Learning}
In recent research on the learning of metrics\cite{chopra2005learning,weinberger2009distance,schroff2015facenet}, contrastive learning has been applied extensively. In recent years, it has been utilized for self-supervised learning (SSL)\cite{oord2018representation,wu2018unsupervised,ye2019unsupervised,tian2020contrastive,he2019momentum,chen2020simple,misra2020self,sermanet2017time,hyvarinen2016unsupervised}, in which it is employed to learn an encoder during the pretext training phase. The goal of contrastive learning approaches in the self-supervised learning environment, in the absence of labels, is to learn a uniform representation for every image in the training set. In order to accomplish that, a contrastive loss assessed upon pairs of feature vectors taken from data augmentations of the image is minimized. Although this fundamental concept is shared by the majority of contrastive learning-based self-supervised learning techniques, various augmentation mechanisms have been presented\cite{wu2018unsupervised,ye2019unsupervised,tian2020contrastive,he2019momentum,chen2020simple,misra2020self}. The most common method of obtaining augmentations is data manipulation (rotation, cropping, random greyscale, and colour jittering)\cite{ye2019unsupervised,chen2020simple}. However, other approaches, such as using depth, surface normals, or other colour channels, have also been proposed\cite{tian2020contrastive}. Utilizing an augmentation dictionary comprised of the embedding vectors derived from the prior epoch\cite{wu2018unsupervised} or one that is produced by passing an image through an encoder that updates momentum\cite{he2019momentum} is an additional method. The wide range of methods used to create augmentations illustrates how crucial it is to use instance sets in contrastive learning that are semantically identical\cite{arora2019theoretical}. This was also empirically investigated\cite{chen2020simple}, which demonstrates that contrastive learning performance is enhanced by more robust data augmentations.\\
The majority of contrastive learning approaches are unable to connect the image instances inside a batch or mine hard negative pairings despite the abundance of augmentation proposals available for self-supervised learning. Although \cite{chen2020simple,noroozi2016unsupervised,misra2016unsupervised,tschannen2020mutual} have discussed the significance of choosing negative pairings, but do not provide a methodical procedure for doing so. Inspired by metric learning's noise contrastive estimation (NCE)\cite{gutmann2010noise} and N-pair\cite{sohn2016improved} loss approaches, contrastive learning inherits the widely recognized challenges of challenging negative mining as documented in this literature\cite{wu2017sampling,schroff2015facenet}. When the dataset grows, the number of potential positive and negative pairings for metric learning algorithms\cite{suh2019stochastic,kumar2017smart} rises substantially (for instance, cubically when the triplet loss is applied\cite{schroff2015facenet}. Drawing negative samples over a noisy distribution that handles all negative samples identically is one way that NCE solves this problem\cite{bose2018adversarial,bose2018compositional}.
\subsection{Adversarial Examples}
To generate adversarial attacks that cause a network to fall short, adversarial instances are generated from clean instances\cite{DBLP:journals/corr/abs-1712-07107,chakraborty2018adversarial,ho2020contrastive}. Numerous supervised learning scenarios have made use of them, such as segmentation\cite{goodfellow2015explaining,kurakin2016adversarial}, object identification, and image categorization\cite{kurakin2016}. Adversarial training involves the practice of training a network through both clean and adversarial instances in order to strengthen its defenses against attacks like this\cite{shafahi2019adversarial}. Moreover, self-supervised learning may be used to strengthen defenses against invisible threats\cite{tramer2020ensemble}. Although adversarial training often works well as a defensive system, the efficacy of clean instance categorization often decreases\cite{kurakin2016adversarial}.\\
Although overfitting to the adversarial instances is often blamed for this consequence\cite{lee2020adversarial}, it is still nearly of a paradox because, in theory, more diverse adversarial instances might enhance standard training\cite{ilyas2019adversarial}, for instance, by helping architectures trained on them generalize more effectively to new data\cite{volpi2018generalizing}. In conclusion, although hostile instances may help with learning, it is yet unknown how to do this. Developing a process known as AdvProp that interprets clean and adversarial instances sampled from distinct domains and employs an alternate set of batch normalization (BN) layers for every domain\cite{xie2019adversarial} has recently achieved headway in this approach.

In our paper, we extend the pair sampling strategy and the objective function of contrastive learning for adversarial defense training. Our modified sampling strategy forms positive pairs from multiple perturbed versions of the same image and negative pairs from multiple perturbed versions of different images. This approach encourages the model to produce similar feature representations for an image and its perturbed versions, using adversarial attack mechanisms like FGSM, PGD, and CW.

\section{Proposed Model}

\subsection{Contrastive Learning}
\begin{figure}[h]
    \centering
    \includegraphics[scale=0.62]{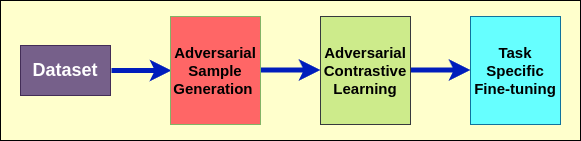}
    \caption{A high-level overview of the proposed framework for adversarial defense with contrastive learning.}
    \label{CL_framework}
\end{figure}

Our innovative framework, grounded in contrastive learning, serves as the cornerstone of our self-supervised training approach, aiming to improve the robustness of deep learning architectures. The contrastive learning loss is crafted to bring similar samples together while pushing dissimilar samples apart, utilizing a contrastive loss function like InfoNCE (normalized cross entropy). This loss is computed by comparing an anchor sample to positive and negative samples, encouraging similar representations for anchor and positive samples while creating distance from negative samples(Eq. \ref{eqloss}). The similarity between samples can be computed using metrics like cosine similarity or euclidean distance. Fig. \ref{CL_framework} provides a high-level overview of the proposed framework for adversarial defense using contrastive learning. 

\begin{align}
L &= -\log\left(\frac{\exp\left(\frac{\text{sim}(x_i, x^+_i)}{\tau}\right)}{\exp\left(\frac{\text{sim}(x_i, x^+_i)}{\tau}\right) + \sum_{j=1}^N \exp\left(\frac{\text{sim}(x_i, x^-_j)}{\tau}\right)}\right) \label{eqloss}
\end{align}

Here, \(\tau\) is a temperature parameter controlling the probability distribution's smoothness, guiding the training process to optimize the backbone’s similarity-based representation learning.

\subsection{Adversarial Sample Generation Strategy}
Through an extensive exploration of various attack techniques, our proposed model showcases resilience against gradient-based attacks during both training and testing stages, regardless of the familiarity of the attacks. Contrastive learning involves a two-step process: representation learning and discriminative learning. In the representation learning phase, a deep neural network, such as a convolutional neural network (CNN), is trained to extract feature representations from input data.

\begin{figure}[H]
    \centering
    \includegraphics[scale=0.46]{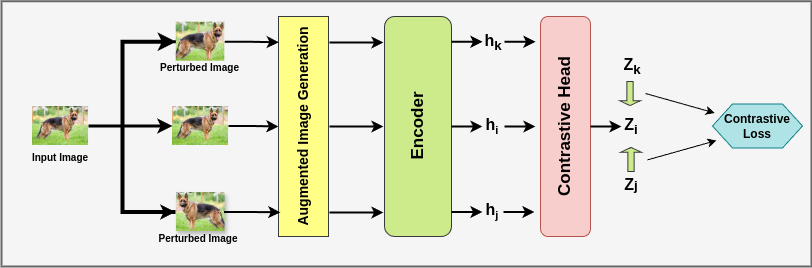}
    \caption{The contrastive learning framework for adversarial defense.}
    \label{Contrastive Learning Training}
\end{figure}

The discriminative learning phase uses a pre-trained encoder to extract features from anchor, positive, and negative samples, with the contrastive loss computed based on these features. Various techniques enhance the learning process, including data augmentations (e.g., random crops, colour jittering) and memory banks storing negative samples for mining hard negatives. Contrastive learning demonstrates promising results in domains like image recognition, object detection, and natural language processing, providing powerful representations without explicit labels.

\subsection{Adversarial Contrastive Training}
A pivotal component in our proposed method is the integration of adversarial contrastive training, leveraging contrastive learning techniques to accentuate the backbone’s similarity-based representation learning. This stage involves the meticulous generation of augmented and perturbed images guided by a contrastive loss function. In the first step of contrastive training, various data augmentation techniques, including random cropping and horizontal flipping, are employed to enhance the backbone’s similarity-based representation learning. The intuition is to encourage the backbone to generate similar feature representations for pairs of original and perturbed augmented images, with a contrastive loss function guiding this objective.\\
The contrastive loss function is calculated by comparing pairs of images, each consisting of an original image and a perturbed image. The loss is the average of contrastive losses between the original image and the PGD-perturbed image and between the original image and the CW-perturbed image. This contrastive loss guides the training process to optimize the backbone’s similarity-based representation learning(Eq. \ref{equ1}).


\begin{align} \label{equ1}
L_{\text{contrastive}} &= -\frac{1}{2} \log\left(\frac{\exp\left(\frac{\text{sim}(x_{\text{orig}}, x_{\text{PGD}})}{\tau}\right)}{\exp\left(\frac{\text{sim}(x_{\text{orig}}, x_{\text{PGD}})}{\tau}\right) +\sum_{j=1}^N \exp\left(\frac{\text{sim}(x_{\text{orig}}, x_j^-)}{\tau}\right) }\right) \nonumber \\
&\quad + \log\left(\frac{\exp\left(\frac{\text{sim}(x_{\text{orig}}, x_{\text{CW}})}{\tau}\right)}{\sum_{j=1}^N \exp\left(\frac{\text{sim}(x_{\text{orig}}, x_j)}{\tau}\right) + \exp\left(\frac{\text{sim}(x_{\text{orig}}, x_{\text{CW}}^-)}{\tau}\right)}\right)
\end{align}

Here, $x_{\text{orig}}$ represents the original image, $x_{\text{PGD}}$ is the PGD-perturbed image, $x_{\text{CW}}$ is the CW-perturbed image, and \(\tau\) is the temperature parameter.

\begin{algorithm}[h]
\caption{Contrastive Learning for Enhanced
Adversarial Defense.}\label{algo1}
\begin{algorithmic}[1]
    \STATE \textbf{Input} D: dataset, $E_{Q}$: encoder, pretrainEpochs: number of epochs for adversarial contrastive training, finetunningEpochs: number of epochs for fine-tunning\\
    \textbf{Initialize} Encoder $E_{Q}$ with random weights 

    //\textbf{Adversarial Contrastive Training (ACT)}
        \FOR{$e = 0$ to pretrainEpochs} 
            \STATE MB $\leftarrow$ Sample mini-batch of size N from D
            \STATE V $\leftarrow$ Obtain corresponding views of each image in MB
            \STATE PGD, CW $\leftarrow$ Obtain two perturbed versions of each image in MB
            \STATE Train Encoder $E_{Q}$ with ACT framework (Fig. \ref{Contrastive Learning Training}) and $L_{\text{contrastive}}$ loss using V, PGD, \& CW.
        \ENDFOR
        
    //\textbf{Task-specific Fine-tunning (TF)}
        \FOR{$e = 0$ to finetunningEpochs} 
            \STATE Transfer pre-trained $E_{Q}$ in TF framework (Fig. \ref{Transfer Learning Training})
            \STATE Fine-tune the framework for the classification task
        \ENDFOR
    \STATE \textbf{Output} TF Framework
\end{algorithmic}
\label{algo}
\end{algorithm}

\subsection{Task-specific Fine-tuning}
\begin{figure}[h]
    \centering
    \includegraphics[scale=0.54]{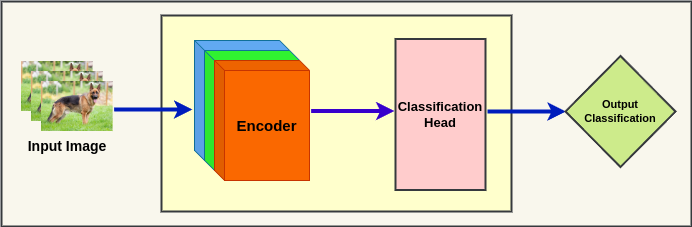}
    \caption{Task-specific fine-tuning framework through transfer learning.}
    \label{Transfer Learning Training}
\end{figure}

The final stage involves task-specific fine-tuning through transfer learning. Extracting the pre-trained backbone from the contrastive training model preserves its weights. Subsequently, a linear layer is introduced and fine-tuned exclusively for the target task, aligning learned features for enhanced performance and generalization ability. In the second step of contrastive learning, known as transfer learning, the trained backbone is adapted to a specific task or dataset. The pre-trained backbone is extracted and frozen to preserve learned features. A linear layer is added and fine-tuned for the target task, mapping learned feature representations to the classes of the CIFAR-10 dataset.
During transfer learning, only the parameters of the added linear layer are trained, while the backbone’s weights remain fixed. This fine-tuning process aligns learned features with the target task, enhancing the model’s performance and generalization ability. Figure \ref{Transfer Learning Training} illustrates the process of task-specific fine-tuning using transfer learning. The detailed steps of the proposed model are summarised in Algorithm \ref{algo}.

\section{Experiments}
The outcomes of our experimental verification, which we carried out to evaluate C-LEAD's effectiveness, are shown in this section. The PyTorch framework was utilised to conduct our research on an NVIDIA DGX Station A100 GPU equipped with 40G memory.

\subsection{Dataset}
We have utilised the benchmark CIFAR-10 dataset for the framework. The data comprises 60000 32x32 color image instances separated into ten classes.
The class labels are airplane, automobile, bird, cat, deer, dog, frog, horse, ship and truck\cite{krizhevsky2009learning}. 
\subsection{Experimental Settings}
For the representation learning problem, we use a 2-layer MLP projection head and a ResNet18-based encoder. 128-dimensional vectors were subsequently formed as a result of this structure. We selected certain hyper-parameters: a temperature coefficient of 0.1, a batch size of 512, a cosine learning rate scheduler, and an SGD optimiser with a momentum of 0.9 and a learning rate of 0.4. A linear layer that generates class probabilities and a ResNet18 base encoder is used for the pseudo-label creation job. The Adam optimiser is used in the model optimisation procedure, with a batch size of 128 and a learning rate of 0.0001.

\subsection{Results and Analysis}
The results of our experiments are divided into two processes: the first process involves attacks that are seen and used during training, while the second process evaluates the performance of the backbone model against baseline attacks both before and after training. 

Table \ref{woaccw} presents a comparison of the clean model without any training against the baseline attacks. Before training, the backbone model exhibits vulnerabilities to these attacks. Subsequently, we compare the performance of the trained model against the same baseline attacks. The results demonstrate an average improvement of 40\% compared to the baseline model. Notably, the attacks used for training, such as PGD and CW, show significant improvements, while the FGSM attack, acting as an unseen attack, still poses a challenge for the trained model.
\\
\begin{table}[h]
\centering
\resizebox{\textwidth}{!}{%
\begin{tabular}{|l|l|l|l|l|l|l|}
\hline
\multirow{2}{*}{\textbf{Model}} & \multirow{2}{*}{\textbf{Clean}} & \multirow{2}{*}{\textbf{Attack}} & \multicolumn{2}{|c|}{\textbf{Accuracy w/o training}} & \multicolumn{2}{|c|}{\textbf{Accuracy w/ training}} \\ 
 & & & $\epsilon = 0.03$ & $\epsilon = 0.06$ & $\epsilon = 0.03$ & $\epsilon = 0.08$ \\ \hline
\multirow{3}{*}{Resnet 18} & \multirow{3}{*}{87.89\%} & FGSM & 18.38\% & 13.60\% & 25.78\% & 24.97\% \\  
& & PGD & 12.60\% & 10.34\% & 31.27\% & 27.61\% \\ 
& & C\&W & 9.80\% & 7.40\% & 21.55\% & 18.80\% \\ \hline
\multirow{3}{*}{Resnet 34} & \multirow{3}{*}{89.96\%} & FGSM & 19.70\% & 16.52\% & 53.23\% & 49.16\% \\  
& & PGD & 14.29\% & 11.40\% & 61.01\% & 58.60\% \\ 
& & C\&W & 17.62\% & 12.95\% & 58.66\% & 51.79\% \\ \hline
\multirow{3}{*}{Resnet 50} & \multirow{3}{*}{93.38\%} & FGSM & 18.47\% & 14.32\% & 55.28\% & 50.86\% \\  
& & PGD & 15.59\% & 11.60\% & 68.67\% & 57.56\% \\ 
& & C\&W & 16.40\% & 13.26\% & 59.85\% & 52.20\% \\ \hline
\end{tabular}%
}
\caption{Table reflecting the accuracy w/o training and w/ training for various models and attacks with $\epsilon$ values.}
\label{woaccw}
\end{table}
\\
Additionally, we examine the impact of the model depth on the results. Deeper models, such as ResNet34 and ResNet50, perform better than ResNet18.\\
However, the effectiveness of the contrastive training approach seems to be less pronounced with ResNet18, resulting in an average improvement of only 10\%-15\%. The limited capacity and feature extraction abilities of ResNet18 might constrain its performance, making the contrastive training benefits less pronounced. Larger models typically capture more complex features, which can lead to more significant improvements with contrastive learning.
To further assess the performance of our proposed approach, we compare it(Table \ref{compare_sota}) with existing adversarial training (AT) defense methods such as PGD-AT\cite{jia2022las} and others\cite{madry2019towards,zhang2020attacks,zhu2022adversarial}. Our proposed approach shows improvements over these baseline AT models, indicating its effectiveness in enhancing robustness against adversarial attacks.
\begin{table}[h]
\centering
\begin{tabular}{|l@{\hspace{1.2cm}}|@{\hspace{1cm}}l@{\hspace{1cm}}|c|c|c|}
\hline
\multirow{2}{*}{\textbf{Adversarial Training Method}}              & \multirow{2}{*}{\textbf{Model}} & \multicolumn{3}{c|}{\textbf{Adversarial Attacks}}\\    
& &\textbf{FGSM} & \textbf{PGD}    & \textbf{CW} \\ \hline
AT \cite{madry2019towards}                  & Resnet18         & \textcolor{green}{\textbf{60.9}}  & \textcolor{green}{\textbf{66.3}} & -              \\ 
PGD-AT (LAS)\cite{jia2022las}        & Resnet34         & -                       & 56.02          & \textcolor{green}{\textbf{53.91}}          \\ 
TRADES \cite{zhang2020attacks}             & Resnet50         & 53.49                   & 63.87 & -              \\ 
LAS-AT\cite{jia2022las}              & Resnet18         & -                       & 61.09          & \textcolor{blue}{\textbf{58.22}}          \\ 
ROCl\cite{zhu2022adversarial}               & Resnet50         & \textcolor{red}{\textbf{67.59}}  & \textcolor{blue}{\textbf{66.76}} & -              \\ \hline
\textbf{Ours}               & Resnet50         & \textcolor{blue}{\textbf{55.28}}                   & \textcolor{red}{\textbf{68.67}}  & \textcolor{red}{\textbf{59.85}} \\ \hline
\end{tabular}%
\caption{Comparison table with other adversarial training defense methods with $\epsilon = 8$. The first, second, and third-best performances are represented in \textcolor{red}{\textbf{red}}, \textcolor{green}{\textbf{green}}, and \textcolor{blue}{\textbf{blue}}, respectively.}
\label{compare_sota}
\end{table}
\\
However, it is important to note that our models, including the proposed approach, may fall short in terms of accuracy when compared to state-of-the-art models that incorporate preprocessing techniques, model modifications, or ensemble learning. The results presented comprehensively compare our models and these adversarial training approaches.
Overall, the results highlight the improvements achieved through our proposed approach while acknowledging the need for further advancements to match the accuracy levels of state-of-the-art models that incorporate advanced techniques like preprocessing and ensemble learning.

\section{Conclusion}\label{sec13}

Our research highlights the effectiveness of using contrastive loss in adversarial training to strengthen model robustness. We found that deeper models like ResNet50 outperformed shallower ones and that smaller batch sizes improved training outcomes. Future directions include using adversarial training as a preprocessing step, exploring image preprocessing techniques, and implementing label smoothing to further enhance model resilience. Additionally, ensemble methods show promise for creating robust models suitable for real-time applications, emphasizing the critical role of contrastive learning in adversarial defense strategies. Furthermore, our experiments indicate that adversarially trained models can effectively resist various types of attacks, demonstrating their potential for deployment in security-critical environments. By continuing to refine these techniques, we can achieve even higher levels of robustness and generalization. The integration of contrastive learning with other advanced training methods offers a pathway to developing state-of-the-art models that are both powerful and secure. As the field evolves, ongoing research and innovation will be crucial in addressing the ever-changing landscape of adversarial threats.

\section{Competing Interests}
On behalf of all authors, the corresponding author states that there is no conflict of interest.

\section{Funding Information}
This research is supported by the Core Research Grant from the Science and Engineering Research Board (SERB), Department of Science and Technology (DST), Government of India, under Grant No. CRG/2020/000651.
\section{Author contribution}
Suklav Ghosh and Sonal Kumar: Methodology, experimentation and manuscript preparation; Arijit Sur: Supervision and manuscript preparation.
\section{Data Availability Statement}
Publicly available CIFAR-10 Dataset.
\section{Research Involving Human and /or Animals}
Not Applicable
\section{Informed Consent}
Not Applicable

\bibliography{sn-bibliography}

\end{document}